\documentclass[letterpaper]{article} 
\usepackage{aaai2026}  
\usepackage{times}  
\usepackage{helvet}  
\usepackage{courier}  
\usepackage[hyphens]{url}  
\usepackage{graphicx} 
\usepackage{fancyhdr} 
\usepackage{etoolbox} 
\makeatletter
\patchcmd{\@maketitle}{\vskip 0.625in minus 0.125in}{\vskip 0.12in minus 0.05in}{}{}
\patchcmd{\@maketitle}{\vskip 0.625in minus 0.125in}{\vskip 0.12in minus 0.05in}{}{}
\makeatother
\usepackage{natbib}  
\usepackage{caption} 
\title{Behavior-Adaptive Conversational Agents: Toward a Fluid Personality Framework}
\author{
    Hasibur Rahman, 
    Smit Desai
}

\affiliations{
    Northeastern University\\
    Boston, MA, USA\\
    rahman.has@northeastern.edu, sm.desai@northeastern.edu
}

\begin{document}
\copyrighttext{Presented at \emph{Bridging AI and Behavior Change}, a Bridge Program
organized at the AAAI Conference on Artificial Intelligence 2026 (AAAI-2026).
\url{https://aaai.org/conference/aaai/aaai-26/bridge-program/}}
\maketitle

\fancypagestyle{bridgehdr}{%
  \fancyhf{}%
  \renewcommand{\headrulewidth}{0pt}%
  \fancyhead[C]{\footnotesize Presented at Bridging AI and Behavior Change, AAAI-2026 Bridge Program}%
}
\setlength{\headheight}{13.6pt}
\setlength{\headsep}{4pt}
\pagestyle{bridgehdr}
\thispagestyle{bridgehdr}

\begin{abstract}
Large language model (LLM)-based conversational agents (CAs) are now ubiquitous, creating new opportunities for AI-mediated behavior change. Their capacity to project nuanced personalities and adopt diverse metaphorical roles raises a design question: \emph{how should an agent’s persona and personality be calibrated to the moment?} Recent evidence suggests (i) \emph{moderate} personality expression outperforms low or high extremes on trust, enjoyment, and intention to adopt in goal-oriented tasks, and (ii) \emph{context-appropriate} metaphors outperform static “one-note” assistants on user experience and uptake. Yet most CAs still fix both persona and style, risking misalignment when dynamics, urgency, and formality vary (e.g., medical information seeking vs.\ fitness coaching vs.\ reflective learning). We propose a \textit{Fluid Personality Framework} that jointly adapts (1) the agent’s \emph{metaphorical persona} (e.g., coach, tutor, librarian/tool) and (2) its \emph{personality expression intensity} (low/medium/high) as a function of task context, user goals and traits, and situational urgency. We sketch the framework and its core design dimensions.
\end{abstract}

\section{Introduction}
LLM-based CAs now span healthcare \cite{furtado_assessing_2025, stade_large_2024, yang_behavioral_2025}, education \cite{vanzo_gpt-4_2025, kestin_ai_2025}, productivity \cite{noy_experimental_2023, brynjolfsson_generative_2025}, and general assistance \cite{chatterji_how_2025, polyportis_longitudinal_2024}. Unlike earlier rule-based systems \cite{chen_survey_2017}, they can express personality via linguistic cues \cite{jiang_personallm_2024, rahman_vibe_2025, serapio-garcia_personality_2025} and embody different personas \cite{chen_persona_2024, wang_rolellm_2024, shao_character-llm_2023}. However, a fixed friendly “assistant” is often applied across contexts \cite{zheng_when_2024}. This can be suboptimal: casual buddy-like behavior may erode credibility in serious domains, while a strictly formal expert can suppress empathy and rapport. Human coaches naturally adjust role and tone; CAs should too. We ask: How should personality be calibrated for different contexts? How should CAs adapt dynamically?

Two lines of recent research highlight both the need for and the feasibility of more dynamic CA personalities. First, work on metaphorical persona design in voice user interfaces shows the limits of static anthropomorphic personas \cite{desai_examining_2024}. \citet{desai_examining_2024} found that framing a voice-based CA as a human-like assistant (e.g., a “doctor” in health or “financial advisor” in finance) can create unrealistic expectations and reinforce stereotypes. User perceptions depended on context: in health, a doctor persona was preferred over a “health encyclopedia,” whereas in finance, the metaphor (human vs. calculator) made little difference. Both human and non-human personas were seen as equally trustworthy and intelligent, suggesting anthropomorphism is not required for user trust. These findings motivate contextual persona adaptation. An agent might adopt a warmer, more human-like demeanor in empathy-driven settings, but a more utilitarian persona when efficiency is key.

Recent work on metaphor-fluid voice-based CAs formalized this idea: allowing an agent to shift personas (e.g., “Genie” for commands, “Star Trek Computer” for information, “Admirer” for social chat, “Search Engine” for errors) increased user enjoyment, likability, and adoption intention compared to a single-persona assistant \cite{desai_toward_2025}. Users valued contextual relevance, while trust and perceived intelligence held steady, showing that persona fluidity can enhance experience without reducing credibility.

Second, a CA’s personality, defined by traits such as openness, conscientiousness, extraversion, emotional stability, and agreeableness, must be balanced in how strongly it is expressed. \citet{rahman_vibe_2025} found an inverted-U relationship between expression intensity and user evaluations: medium expression outperformed both low and high extremes across trust, intelligence, and enjoyment. Likewise, \citet{volkel_examining_2021} showed that adjusting Agreeableness influences warmth, while personality mirroring can improve user perceptions and self-disclosure \cite{shumanov_making_2021, gnewuch_effect_2020}. Yet \citet{spagnolli_similarity_2025} found no such effect in a health setting, underscoring the need for context-aware personality design.

To date, these two lines of work have advanced largely in isolation. Research on metaphorical persona asks which role an agent should play, while research on personality expression asks how strongly its traits should come through. In practice a user encounters both at once, and the fitting persona and the fitting level of expression each shift with the task, the domain, and the moment. Adapting one dimension while holding the other fixed risks a mismatch, for example a well-chosen coach persona that still overwhelms the user with excessive enthusiasm. We therefore see a need for a single framework that adapts persona and expression jointly, so that an agent can stay suited to a user’s changing goals and contexts. We propose a \emph{Fluid Personality Framework} that enables CAs to dynamically adapt along two orthogonal dimensions: Metaphorical Persona (the role or archetype the agent embodies) and Personality Expression Intensity (the strength and explicitness of personality trait expression).

\section{Related Work}
\subsection{Metaphorical Persona Design}
Metaphors shape users’ partner models and expectations \cite{carroll_chapter_1988, carroll_metaphor_1982}. Commercial CAs default to an anthropomorphic “assistant” \cite{sciuto_hey_2018, ouyang_training_2022, zheng_when_2024}, which can miscalibrate trust \cite{luger_like_2016}. Users already reframe roles across contexts \cite{desai_metaphors_2023, pradhan_phantom_2019}. Empirically, metaphor effects are domain-dependent and shaped by perceived formality and user preferences \cite{jung_great_2022, chin_like_2024, braun_at_2019}; ethical pitfalls of fixed personas are well-documented \cite{brahnam_dressing_2011, mcmillan_leaving_2021}. Metaphor-fluid designs respond by adapting presentational role to task phase and context \cite{desai_metaphors_2023, desai_toward_2025}.

\subsection{Personality in LLM-based CAs}
People apply social rules to machines \cite{nass_computers_1994}; the Big Five \cite{john_big_1991, john_big-five_1999} has been used to script extremes (e.g., Extraversion/Agreeableness) \cite{zhou_trusting_2019, volkel_user_2022, ruane_user_2020, volkel_examining_2021}. LLMs enable prompt-based multi-trait control \cite{ramirez_controlling_2023, jiang_evaluating_2023, jiang_personallm_2024, serapio-garcia_personality_2025}, but sustained mid-range control and role fidelity remain challenging \cite{kovacevic_chatbots_2024}. Alignment between user and agent traits yields mixed results \cite{shumanov_making_2021, gnewuch_effect_2020, isbister_consistency_2000, spagnolli_similarity_2025}; “how much” personality is shown can matter as much as “which” traits \cite{volkel_examining_2021}. In goal-oriented LLM interactions, \emph{medium} expression frequently wins \cite{rahman_vibe_2025}.

\section{Fluid Personality Framework}

We propose a Fluid Personality Framework that unifies metaphor-fluid persona design and adaptive personality expression into a single model for CAs. The agent continuously evaluates two kinds of input. The first is contextual factors such as task type, domain, and urgency. The second is user-specific factors such as personality traits and interaction history. Based on these inputs, the agent dynamically configures its Persona Module and Personality Module before each dialogue turn.

\subsection{Persona Module (Metaphor Adaptor)}

This module selects the most suitable persona or role metaphor for the situation. The persona defines who the agent is “being” (e.g., coach, friend, expert, tool). Rather than a fixed identity, the agent maintains a portfolio of personas it can switch between fluidly. For example, it might begin as a “Planner” during goal-setting (organized, pragmatic), switch to a “Cheerleader” to celebrate progress (upbeat, supportive), and become a “Tutor” when explaining concepts (analytical, patient). Context cues or user needs trigger these shifts. Inspired by \citet{desai_toward_2025}, this approach extends metaphor-fluid design to include non-human metaphors (e.g., a “Library” or “Guide” persona in information-heavy dialogues). Implementation may use prompt engineering (e.g., “You are now speaking as a supportive friend…”) or dialogue logic. Transitions use subtle linguistic bridging to maintain coherence, so the user perceives the agent as responsive rather than static, sometimes a coach and sometimes a companion.

\subsection{Personality Module (Trait Intensity Modulator)}

This module modulates how the agent expresses itself within the chosen persona by adjusting tone, affect, formality, and stylistic intensity. Building on the Trait Modulation Keys approach \cite{rahman_vibe_2025}, it selects appropriate levels of traits such as enthusiasm (Extraversion), friendliness (Agreeableness), and conscientiousness. Modulation depends on both context and user. For urgent tasks (e.g., medical reminders), the agent may reduce “agreeable chit-chat,” adopting a concise, serious tone (lower agreeableness, higher conscientiousness). For long-term behavior change, it may increase empathy and encouragement to sustain motivation. The module also aligns with the user’s profile, for example lowering extraversion to avoid verbosity for introverted users. It essentially answers, “Given my persona, how expressive should I be right now to serve the user best?” This dynamic tuning keeps the agent within the “Goldilocks zone” identified by \citet{rahman_vibe_2025}.

\section{Conclusion}
Moving from static to adaptive personality, the Fluid Personality Framework couples metaphorical persona selection with trait-intensity modulation. By tuning role and expression to context, user traits, and task demands, CAs can improve user experience and effectiveness in behavior-change settings.

\bibliography{aaai2026}
\end{document}